\documentclass[10pt, a4paper]{article}

\usepackage[final]{lrec2026}
\usepackage{amsmath}
\usepackage{amssymb}
\usepackage{booktabs}
\usepackage{multirow}
\usepackage{graphicx}
\usepackage{microtype}
\usepackage[nolist]{acronym}
\title{LabelFusion: Fusing Large Language Models with Transformer Encoders for Robust Financial News Classification}

\name{
{\large\bf Michael Schlee$^{1}$, Christoph Weisser$^{1}$, Timo Kivim\"aki$^{3}$}\\
{\large\bf Melchizedek Mashiku$^{4}$, Benjamin Säfken$^{5}$}
}

\address{
$^{1}$Centre for Statistics, Georg-August-Universit\"at G\"ottingen, Germany\\
$^{2}$Hochschule Bielefeld (HSBI) - University of Applied Sciences and Arts, Bielefeld, Germany \\
$^{3}$Department of Politics and International Studies, University of Bath, Bath, UK\\
$^{4}$Tanaq Management Services LLC, Contracting Agency to the Division of Viral Diseases\\
Centers for Disease Control and Prevention, Chamblee, Georgia, USA\\
$^{5}$Institute of Mathematics, Clausthal University of Technology, Clausthal-Zellerfeld, Germany\\
michael.schlee@uni-goettingen.de, christoph.weisser@hsbi.de, t.kivimaki@bath.ac.uk \\
melchizedek.mashiku@tanaq.com,
benjamin.saefken@tu-clausthal.de
}

\abstract{
Financial news plays a central role in shaping investor sentiment and short-term dynamics in commodity markets. Many downstream financial applications—such as commodity price prediction or sentiment modeling—therefore rely on the ability to automatically identify news articles that are relevant to specific assets. However, obtaining large labeled corpora for financial text classification tasks is costly, and transformer-based classifiers such as RoBERTa often degrade significantly in low-data regimes. Our results show that appropriately prompted out-of-the-box \acp{LLM} achieve strong performance even in low-data regimes. Furthermore, we propose LabelFusion, a hybrid architecture that combines the output of a prompt-engineered \ac{LLM} with contextual embeddings produced by a fine-tuned RoBERTa encoder through a lightweight \ac{MLP} voting layer. Evaluated on a ten-class multi-label subset of the Reuters-21578 corpus, LabelFusion achieves a macro F1 score of 96.0\% and an accuracy of 92.3\% when trained on the full dataset, outperforming both standalone RoBERTa (F1 94.6\%) and the standalone \ac{LLM} (F1 93.9\%). In low- to mid-data regimes, however, the \ac{LLM} alone proves surprisingly competitive, achieving an F1 score of 75.9\% even in a zero-shot setting and consistently outperforming LabelFusion until approximately 80\% of the training data is available. These results suggest that \ac{LLM}-only prompting represents the preferred strategy under annotation constraints, whereas LabelFusion becomes the most effective solution once sufficient labeled data is available to train the encoder component. The code is available in an anonymized repository.
\\ \newline
}

\begin{document}

\maketitleabstract

\begin{acronym}
\acro{LLM}{Large Language Model}
\acro{MLP}{Multilayer Perceptron}
\acro{NLP}{Natural Language Processing}
\acro{TF-IDF}{Term Frequency--Inverse Document Frequency}
\acro{SVM}{Support Vector Machine}
\acro{KNN}{$k$-Nearest Neighbors}
\acro{LSTM}{Long Short-Term Memory}
\acro{GCN}{Graph Convolutional Network}
\acro{PET}{Pattern-Exploiting Training}
\acro{CLS}{Classification Token}
\acro{R21578}{Reuters-21578}
\acro{LLM}{Large Language Model}
\acro{MLP}{Multilayer Perceptron}
\acro{TFIDF}{Term Frequency–Inverse Document Frequency}
\acro{LR}{Logistic Regression}
\end{acronym}

% ---------------------------------------------------------------
\section{Introduction}
% ---------------------------------------------------------------

Emotional factors play an substantial role in the decision-making processes of both institutional and individual investors. Such emotional signals exert a measurable influence on individual commodity returns, including assets such as oil or gold, and can even enable short-term predictive insights \cite{sinha2020impactnews}.\\
News articles on specific commodities can act as a primary source of these emotional signals; consequently, financial news represents a valuable resource for identifying market sentiment and potentially anticipating subsequent developments in commodity prices.\\
Transformer-based models such as FinBERT are commonly employed to extract sentiment from financial news by fine-tuning pre-trained language models on labeled financial text corpora \cite{araci2019finbert}. Different model families integrate these sentiment scores in different ways. Hybrid \ac{LSTM} models use transformer-extracted sentiment as additional input features that vary in time along with price data \cite{chae2023sp500,yang2022lasso,nabipour2026federated}. Transformer-based approaches incorporate sentiment through attention mechanisms that weight emotional signals according to their relevance to price movements \cite{chen2024dual}. \\
An important preliminary step in news-based financial prediction tasks, such as forecasting sentiment in news to predict future gold or oil prices, is the filtering of relevant articles. Only news that is directly related to the specific commodity provides meaningful and informative training data for subsequent modeling tasks. This data filtering step is therefore as critical as the sentiment classification task itself. The extremely large volume of available financial news makes manual selection infeasible, creating a clear need for automated text classification methods capable of identifying documents that are relevant to specific commodities or financial assets.\\
LabelFusion addresses this gap by combining state-of-the-art \acp{LLM} with a fine-tuned RoBERTa encoder through an intelligent \ac{MLP}-based voting mechanism. This approach enables effective multi-label text classification while reducing dependence on extensive manually labeled training data. Our contributions are as follows:
\begin{enumerate}
    \item We show that out-of-the-box \acp{LLM}, when used with appropriate prompting, achieve high classification accuracy even when training data is scarce.
     \item We introduce a hybrid fusion model that combines \ac{LLM} predictions with RoBERTa embeddings via a trainable \ac{MLP}. With sufficient training data (80\%--100\%), the approach improves the F1 score from 0.939 to 0.960 compared to a standalone fine-tuned RoBERTa model.
    \item We present \textit{LabelFusion}, a user-friendly software package that facilitates the integration and deployment of fusion-based classification models \cite{labelfusion2025}.
\end{enumerate}

% ---------------------------------------------------------------
\section{Related Work}
% ---------------------------------------------------------------

The Reuters-21578 corpus \cite{lewis1997reuters} has served as the standard benchmark
for multi-label financial news categorization for decades, with classical classifiers
such as \ac{SVM}, \ac{KNN}, and Rocchio establishing strong baselines under severe label
imbalance \cite{debole2005analysis}. More recent work has applied \acp{GCN} that propagate
semantic information across document, word, and label nodes \cite{sgcn2024kbs}, as well
as attention-based architectures that explicitly model label correlations
\cite{yuan2024lacn,liam2024npl}. Despite their strong empirical results, all of these
methods depend on substantial labeled training data and do not incorporate the broad
language understanding that \acp{LLM} provide.\\
\citet{brown2020gpt3} demonstrated that GPT-3 achieves competitive performance across
a wide range of \ac{NLP} benchmarks through in-context few-shot prompting without any
gradient updates, an intuition formalized by \citet{schick2021pet} through
cloze-style \ac{PET} and refined by \citet{gao2021lmbff} via automatic prompt and
verbalizer search. More recent evaluations confirm that instruction-tuned \acp{LLM}
are effective zero-shot classifiers \cite{wang2023zeroshot}, although fine-tuned
encoder models retain a competitive advantage when sufficient labeled data is
available \cite{chae2025llmclassification}.\\
Ensemble and fusion approaches that combine predictions from multiple pre-trained
models consistently outperform individual classifiers \cite{abburi2023ensemble},
and more tightly integrated architectures show that encoder embeddings and
\ac{LLM}-derived features contribute complementary information
\cite{koloski2024automl,gwak2025layerfusion}. To our knowledge, no prior work has
explicitly fused prompt-based \ac{LLM} predictions with fine-tuned encoder
representations for multi-label financial news classification. LabelFusion
addresses this gap by combining both sources through a trainable \ac{MLP}
voting layer, enabling robust performance across the full spectrum of labeled
data availability.

% ---------------------------------------------------------------
\section{Model Architecture}
% ---------------------------------------------------------------

Let $\mathbf{x}$ denote an input text to be assigned labels from a predefined label set $\mathcal{Y} = \{1, \ldots, K\}$, where $K$ denotes the total number of possible categories. The task is multi-label classification, meaning that multiple labels may be associated with a single input text. LabelFusion combines two complementary components — a prompt-based \ac{LLM} and a fine-tuned RoBERTa encoder — whose outputs are fused by a trainable \ac{MLP} voting layer.

\subsection{Prompt-Based LLM Component}

The input text $\mathbf{x}$ is inserted into a prompt template
\[
  p(\mathbf{x}) = \mathcal{T}(\mathbf{x},\, \mathcal{Y},\, \mathcal{E}),
\]
where $\mathcal{T}(\cdot)$ denotes the prompt template, $\mathcal{Y}$ represents the list of predefined labels, and $\mathcal{E}$ denotes an optional set of demonstration examples. Three prompt regimes are considered: \textit{zero-shot} ($\mathcal{E} = \emptyset$), \textit{one-shot} ($|\mathcal{E}|=1$), and \textit{few-shot} ($|\mathcal{E}|>1$). The prompt is processed by a \ac{LLM}
\[
f_{\mathrm{LLM}}: p(\mathbf{x}) \mapsto \mathbf{z},
\]
where $\mathbf{z} \in \{0,1\}^K$ is a binary prediction vector. Each element $z_k$ is defined as $z_k = 1$ if label $k$ is predicted to be present and $z_k = 0$ otherwise.

\subsection{RoBERTa Representation Component}

The same input text is independently encoded by a fine-tuned RoBERTa model:
\[
  h = f_{\mathrm{RB}}(\mathbf{x}) \in \mathbb{R}^{768},
\]
where $h$ corresponds to the contextual embedding of the \texttt{[CLS]} token, capturing rich task-specific semantic information from the input text.

\subsection{Feature Fusion and Prediction}

\begin{figure}[t]
  \centering
  \includegraphics[scale=0.35]{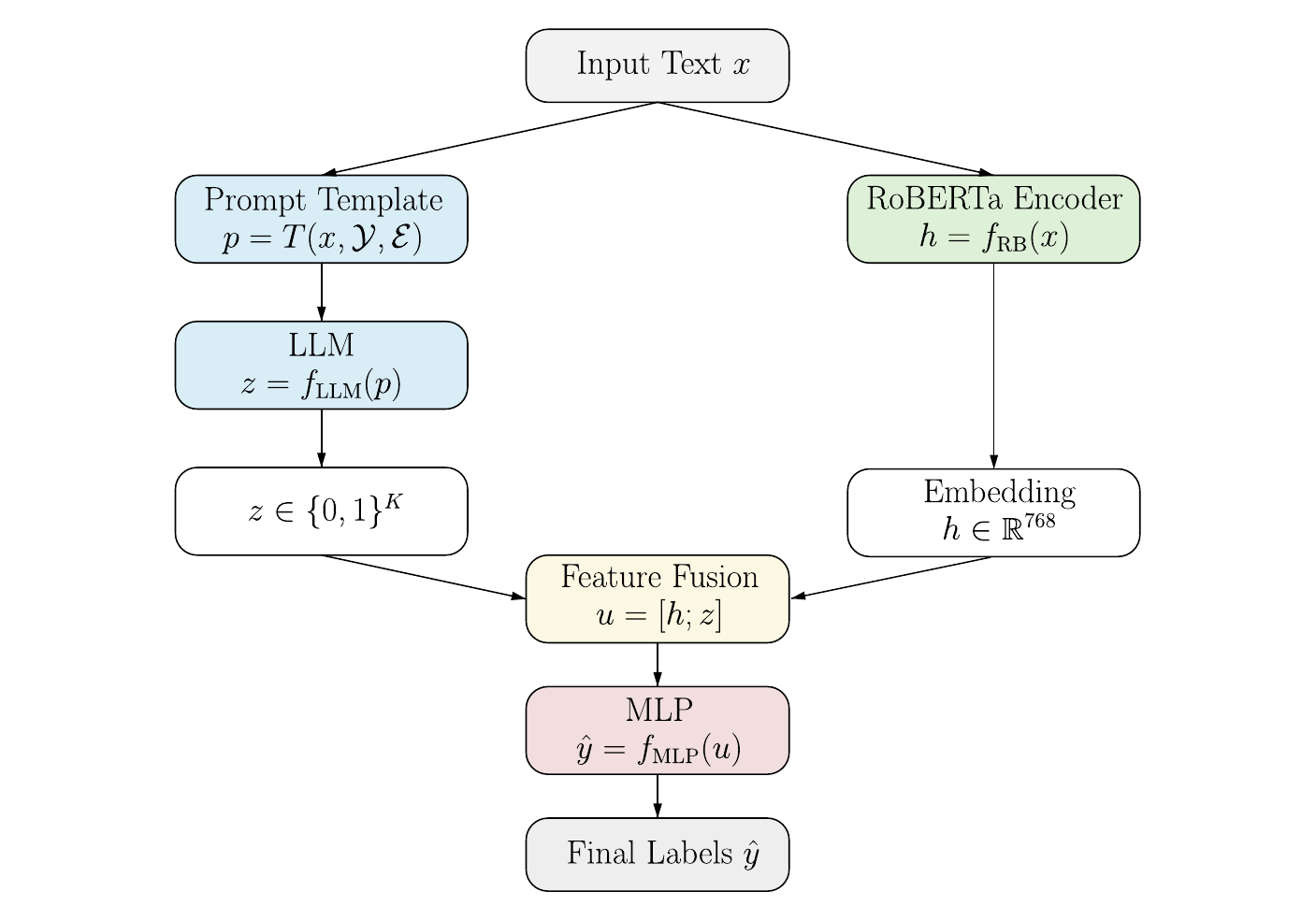}
\caption{Architecture of \textit{LabelFusion}. An Input text $x$ is processed along two parallel branches: a prompt template $T(x, \mathcal{Y}, \mathcal{E})$ feeds a \ac{LLM} to produce a binary label vector $z \in \{0,1\}^K$, while a RoBERTa encoder produces a contextual embedding $h \in \mathbb{R}^{768}$. The two representations are concatenated into a joint feature vector $u = [h\,;\,z]$, which is passed to a trainable \ac{MLP} that outputs the final label predictions $\hat{y}$.}

  \label{fig:architecture}
\end{figure}

The outputs of both components are concatenated to form a fused representation:
\[
  u = [h\,;\,z] \in \mathbb{R}^{768+K}.
\]
This fused vector combines the dense contextual embedding from RoBERTa with the discrete label predictions from the \ac{LLM}, and is passed through a \ac{MLP} voting model:
\[
  \hat{\mathbf{y}} = f_{\mathrm{MLP}}(u) \in [0,1]^K,
\]
where each element $\hat{y}_k$ represents the predicted probability of label $k$. The architecture is illustrated in Figure~\ref{fig:architecture}.

% ---------------------------------------------------------------
\section{Experimental Setup}
% ---------------------------------------------------------------

\subsection{Dataset}

The dataset was constructed from the \ac{R21578} corpus, which contains
approximately 12,902 documents spanning around 135 topic categories with a highly
skewed label distribution. To obtain a manageable and well-populated benchmark, only
the ten most frequent topics were retained: \textit{earn}, \textit{acq},
\textit{money-fx}, \textit{grain}, \textit{crude}, \textit{trade}, \textit{interest},
\textit{ship}, \textit{wheat}, and \textit{corn}. Documents containing none of the
selected topics were excluded, yielding 9,034 documents. Each article was represented
using its full raw text, and a multi-label binary target vector of length ten was
created for each document. Despite the filtering, the dataset remains challenging:
label frequencies range from 3,964 documents for \textit{earn} to only 237 for
\textit{corn}, and 9.2\% of documents carry two or more topic labels simultaneously.\\
The corpus provides an official benchmark split into training and test data, which
was preserved to ensure comparability with prior work. The test set (2,545 documents)
was kept intact from the original ModApte split. From the training portion, a
validation set of 647 documents (10\%) was carved out using multilabel stratified
sampling to maintain a consistent topic distribution across splits, leaving 5,842
documents for training. This procedure yielded three subsets: a training set used for
model learning, a validation set for hyperparameter tuning and model selection, and a
held-out test set used exclusively for the final evaluation of model performance.

\subsection{Experimental Setup}

To evaluate the proposed LabelFusion architecture, we conduct experiments with varying amounts of labeled training data. We construct training subsets comprising 5\%, 10\%, 20\%, 40\%, 60\%, 80\%, and 100\% of the available labeled dataset (5,842 samples). Each subset is used to train the supervised components of the model (RoBERTa and the fusion \ac{MLP}), while the \ac{LLM} branch remains prompt-based. For each subset, the respective proportion of training data is incorporated into the prompt template, which is then sent to the \ac{LLM}.

We compare LabelFusion against several baselines:

\begin{itemize}
\item RoBERTa, a transformer encoder used as a standalone classifier, is fine-tuned on the respective proportion of the training data.
\item GPT-5-nano receives training data subsets embedded in our prompt template. Additionally, we evaluate the performance of the out-of-the-box model in a zero-shot setting, where no training data is used and the prompt template is not applied. Furthermore, we assess classification performance in ultra-low data scenarios, where only a single example is included in the prompt template.
\item \ac{TFIDF} + \ac{LR}, a classical linear baseline trained on the full dataset.
\end{itemize}

For each configuration, we report Accuracy, F1 score, Precision, and Recall. These metrics allow us to evaluate both overall prediction quality and the balance between false positives and false negatives.

% ---------------------------------------------------------------
\section{Results \& Discussion}
% ---------------------------------------------------------------

Table~\ref{tab:results} reports macro F1, accuracy, precision, and recall across all
training data fractions and models. The classical \ac{TFIDF}+\ac{LR} baseline achieves
80.6\% accuracy and a macro F1 of 68.2\%, with high precision (95.4\%) but low recall
(56.9\%), reflecting conservative prediction behavior.

\begin{table}[t]
\centering
\label{tab:results}
\resizebox{1\linewidth}{!}{
\begin{tabular}{llcccc}
\toprule
Data & Model & F1 & Acc. & Prec. & Rec. \\
\midrule
0-shot       & GPT-5-nano  & 75.9 & 83.4 & 89.2 & 70.5 \\
1-shot       & GPT-5-nano  & \textbf{76.1} & 84.1 & 92.0 & 68.0 \\
\midrule
5\% (292)    & Fusion      & 71.7 & 70.6 & 72.0 & 71.5 \\
5\% (292)    & RoBERTa     & 37.2 & 0.0 & 27.6 & 71.3 \\
5\% (292)    & GPT-5-nano  & \textbf{93.0} & 88.1 & 95.2 & 91.7 \\
\midrule
10\% (584)   & Fusion      & 67.1 & 67.0 & 67.2 & 67.1 \\
10\% (584)   & RoBERTa     & 41.7 & 40.0 & 32.1 & 61.6 \\
10\% (584)   & GPT-5-nano  & \textbf{93.8} & 88.5 & 96.2 & 92.6 \\
\midrule
20\% (1168)  & Fusion      & 75.2 & 72.0 & 76.9 & 74.5 \\
20\% (1168)  & RoBERTa     & 53.4 & 67.3 & 46.5 & 64.3 \\
20\% (1168)  & GPT-5-nano  & \textbf{92.8} & 88.6 & 95.1 & 92.3 \\
\midrule
40\% (2336)  & Fusion      & 88.6 & 83.6 & 89.3 & 88.9 \\
40\% (2336)  & RoBERTa     & 83.6 & 82.0 & 85.8 & 85.0 \\
40\% (2336)  & GPT-5-nano  & \textbf{93.1} & 87.9 & 95.2 & 91.7 \\
\midrule
60\% (3505)  & Fusion      & 93.2 & 85.5 & 92.9 & 95.0 \\
60\% (3505)  & RoBERTa     & 90.7 & 83.4 & 90.6 & 94.5 \\
60\% (3505)  & GPT-5-nano  & \textbf{93.8} & 88.4 & 95.9 & 92.4 \\
\midrule
80\% (4673)  & Fusion      & \textbf{95.4} & 90.2 & 95.4 & 96.5 \\
80\% (4673)  & RoBERTa     & 94.3 & 88.8 & 93.0 & 96.6 \\
80\% (4673)  & GPT-5-nano  & 93.4 & 88.0 & 95.1 & 91.8 \\
\midrule
100\% (5842) & Fusion      & \textbf{96.0} & 92.3 & 96.7 & 96.1 \\
100\% (5842) & RoBERTa     & 94.6 & 89.0 & 93.2 & 96.6 \\
100\% (5842) & GPT-5-nano  & 93.9 & 88.9 & 96.3 & 92.7 \\
\midrule\midrule
100\% (5842) & \ac{TFIDF}+\ac{LR}   & 68.2 & 80.6 & 95.4 & 56.9 \\
\bottomrule
\end{tabular}
}
\caption{Performance comparison of LabelFusion, standalone RoBERTa, GPT-5-nano, and a \ac{TFIDF} + \ac{LR} baseline across varying amounts of labeled training data. GPT-5-nano is evaluated in one-shot mode throughout; the zero-shot row reports its performance without any training data. \ac{TFIDF} + \ac{LR} is trained on the full dataset and included as a classical reference. All metrics are macro-averaged and reported in percent (\%) on the Reuters-21578 test set.}
\end{table}
Several clear patterns emerge from Table~\ref{tab:results}. Even in zero-shot mode, the standalone \ac{LLM} already achieves a macro F1 of 75.9\%, exceeding the \ac{TFIDF}+\ac{LR} baseline (68.2\%) by a large margin, likely due to the broad general knowledge and natural language understanding acquired during pretraining. Moving to one-shot mode yields only a marginal gain (76.1\%), suggesting that the \ac{LLM} requires more than a single demonstration to benefit meaningfully from in-context examples.
\\
In the ultra-low data regime, the standalone RoBERTa classifier struggles severely. With only 5\% of the training data (292 documents), it achieves an accuracy of 0.0\% and a macro F1 score of 37.2\%, indicating that the model has not yet learned a reliable decision boundary. With 10\% of the data (584 documents), performance improves to an F1 score of 41.7\%, but remains far below the \ac{LLM}. In contrast, the \ac{LLM} achieves an F1 score of 93.0\% at 5\% and 93.8\% at 10\%, demonstrating that it is a highly effective standalone solution in low-data settings. LabelFusion in these regimes achieves F1 scores of 71.7\% and 67.1\% at 5\% and 10\%, which are better than standalone RoBERTa but inferior to the \ac{LLM} alone. We interpret this as a consequence of the poorly trained RoBERTa component, which appears to introduce noise into the voting process and thereby degrades the overall fusion performance relative to the \ac{LLM} in isolation.\\
In the low- to mid-data regime spanning 5\% to 60\% of the training data, the \ac{LLM} consistently outperforms both standalone RoBERTa and LabelFusion with respect to macro F1 score. We further observe that using only 5\% of the training data is sufficient for the \ac{LLM} to achieve one of the highest F1 scores in the overall classification task. This finding highlights the potential of \acp{LLM} as a standalone approach. In the low- to mid-data regime, LabelFusion is still hampered by noise introduced by the insufficiently fine-tuned RoBERTa encoder, whose weak task-specific signal interferes with the more reliable \ac{LLM} predictions in the voting layer. Nevertheless, a clear upward trend can be observed: as the proportion of training data increases, the RoBERTa component begins to contribute increasingly useful representations, leading to a steady improvement in LabelFusion’s overall accuracy.\\
From 80\% of the training data onward, this trend reaches a tipping point and the situation reverses decisively. LabelFusion surpasses both standalone methods, reaching an accuracy of 90.2\% and a macro F1 score of 95.4\% at 80\%, and achieving the best overall performance with 92.3\% accuracy and a macro F1 score of 96.0\% when trained on the full dataset. These results confirm that once the RoBERTa component is sufficiently trained, the fusion mechanism effectively combines the task-specific discriminative power of the transformer with the broad language understanding of the \ac{LLM}, resulting in a model that is more robust than either component in isolation.

% ---------------------------------------------------------------
\section{Conclusion \& Future Work}
% ---------------------------------------------------------------

We introduced LabelFusion, a hybrid architecture that fuses the logits of a
prompt-engineered \ac{LLM} with embeddings from a fine-tuned RoBERTa encoder via a
trainable \ac{MLP} voting layer for multi-label financial news classification.
Experiments on the Reuters-21578 corpus show that the two model families are
complementary and that the optimal strategy depends on the available annotation
budget: a carefully prompted \ac{LLM} constitutes the strongest standalone solution
in low- to mid-data regimes, while LabelFusion becomes the preferred choice once
sufficient labeled data is available—approximately from 80\% of the training data
onward—where the fine-tuned encoder provides reliable task-specific signals that
the fusion layer can exploit effectively.\\
Future work will explore the integration of additional modalities into LabelFusion. The architecture is designed for the seamless integration of further input sources. In particular, we hypothesise that recent stock price time series of corresponding commodities significantly influence the probability that the commodity appears in financial news. The next logical step would therefore be the integration of historical commodity prices as an additional input modality. These time series could be processed by time-series transformer architectures, which are well suited for extracting short-term temporal patterns through attention mechanisms, and could thereby further improve the macro F1 score of the fusion model.

\section{Acknowledgments}
The work presented in this paper was conducted independently by the author Melchizedek Mashiku and is not affiliated with Tanaq Management Services LLC, Contracting Agency to the Division of Viral Diseases, Centers for Disease Control and Prevention, Chamblee, Georgia, USA.
% ---------------------------------------------------------------
\newpage
\section{Bibliographical References}
\bibliographystyle{lrec2026-natbib}
\bibliography{labelfusion}

@article{sinha2020impactnews,
  author    = {Sinha, Anand and Shastri, Ravi},
  title     = {Impact of News Sentiment on Commodity Returns},
  journal   = {Journal of Behavioral Finance},
  year      = {2020},
  volume    = {21},
  number    = {3},
  pages     = {310--325}
}

@misc{araci2019finbert,
  author        = {Araci, Dogu},
  title         = {{FinBERT}: Financial Sentiment Analysis with Pre-trained Language Models},
  year          = {2019},
  eprint        = {1908.10063},
  archivePrefix = {arXiv},
  primaryClass  = {cs.CL},
  url           = {https://arxiv.org/abs/1908.10063}
}

@article{sgcn2024kbs,
  author    = {Dingkun Zeng and Erxue Zha and Jian Kuang and Yong Shen},
  title     = {Multi-Label Text Classification Based on Semantic-Sensitive Graph Convolutional Network},
  journal   = {Knowledge-Based Systems},
  volume    = {282},
  pages     = {111107},
  year      = {2024},
  publisher = {Elsevier},
  url       = {https://doi.org/10.1016/j.knosys.2023.111107}
}

@inproceedings{yuan2024lacn,
  author    = {Hao Yuan and Weiguang Han and Yucheng Li},
  title     = {{LACN}: Label-Aware Co-training Network for Multi-Label Text Classification},
  booktitle = {Proceedings of the 2024 Joint International Conference on Computational Linguistics, Language Resources and Evaluation (LREC-COLING)},
  pages     = {3412--3422},
  year      = {2024},
  publisher = {ELRA and ICCL},
  url       = {https://aclanthology.org/2024.lrec-main.303}
}

@article{liam2024npl,
  author    = {Yilin Ma and Xin Gao and Bowen Yang},
  title     = {{LIAM}: Label-Interaction Aware Model for Multi-Label Text Classification},
  journal   = {Neurocomputing},
  volume    = {580},
  pages     = {127139},
  year      = {2024},
  publisher = {Elsevier},
  url       = {https://doi.org/10.1016/j.neucom.2024.127139}
}

@article{chae2023sp500,
  author  = {Chae, Seung Chan and Choi, Sun-Yong},
  title   = {Forecasting the {S\&P} 500 Index Using Mathematical-Based Sentiment Analysis and Deep Learning Models},
  journal = {Axioms},
  year    = {2023},
  volume  = {12},
  number  = {9},
  pages   = {835},
  doi     = {10.3390/axioms12090835}
}

@article{yang2022lasso,
  author  = {Yang, Jianfeng and Wang, Yufei and Li, Xiang},
  title   = {Prediction of Stock Price Direction Using the {LASSO-LSTM} Model Combining Technical Indicators and Financial Sentiment Analysis},
  journal = {PeerJ Computer Science},
  year    = {2022},
  volume  = {8},
  pages   = {e1148},
  doi     = {10.7717/peerj-cs.1148}
}

@inproceedings{nabipour2026federated,
  author    = {Morteza Nabipour and Pejman Naeem and Hamed Jabani},
  title     = {Federated Learning for Financial Sentiment-Augmented Price Prediction},
  booktitle = {Proceedings of the 2026 International Conference on Machine Learning and Applications (ICMLA)},
  year      = {2026},
  publisher = {IEEE}
}

@article{chen2024dual,
  author    = {Yilin Chen and Xiaolong Li and Yonghong Hu},
  title     = {Dual-Attention Transformer for Financial News Sentiment and Stock Price Prediction},
  journal   = {Expert Systems with Applications},
  volume    = {238},
  pages     = {121134},
  year      = {2024},
  publisher = {Elsevier},
  url       = {https://doi.org/10.1016/j.eswa.2023.121134}
}

@inproceedings{brown2020gpt3,
  author    = {Brown, Tom B. and others},
  title     = {Language Models are Few-Shot Learners},
  booktitle = {Advances in Neural Information Processing Systems},
  volume    = {33},
  pages     = {1877--1901},
  year      = {2020}
}

@inproceedings{schick2021pet,
  author    = {Schick, Timo and Sch{\"u}tze, Hinrich},
  title     = {Exploiting Cloze-Questions for Few-Shot Text Classification},
  booktitle = {EACL},
  pages     = {255--269},
  year      = {2021}
}

@inproceedings{gao2021lmbff,
  author    = {Gao, Tianyu and Fisch, Adam and Chen, Danqi},
  title     = {Making Pre-trained Language Models Better Few-Shot Learners},
  booktitle = {ACL},
  pages     = {3816--3830},
  year      = {2021}
}

@article{wang2023zeroshot,
  author  = {Wang, Zhiqiang and Pang, Yiran and Lin, Yanbin},
  title   = {Large Language Models are Zero-Shot Text Classifiers},
  journal = {arXiv preprint arXiv:2312.01044},
  year    = {2023}
}

@article{chae2025llmclassification,
  author    = {Chae, Youngjin and Davidson, Thomas},
  title     = {Large Language Models for Text Classification},
  journal   = {Sociological Methods \& Research},
  year      = {2025}
}

@article{abburi2023ensemble,
  author  = {Abburi, Harika and others},
  title   = {Generative AI Text Classification Using Ensemble LLM Approaches},
  journal = {arXiv preprint arXiv:2309.07755},
  year    = {2023}
}

@inproceedings{koloski2024automl,
  author    = {Koloski, Boshko and Pollak, Senja and Navigli, Roberto and {\v{S}}krlj, Bla{\v{z}}},
  title     = {AutoML-Guided Fusion of Entity and LLM-Based Representations},
  booktitle = {ICONIP},
  pages     = {89--102},
  year      = {2024}
}

@article{gwak2025layerfusion,
  author  = {Gwak, Jungwoo and Jung, Yoonsang},
  title   = {Layer-Aware Embedding Fusion for LLMs},
  journal = {arXiv preprint arXiv:2504.05764},
  year    = {2025}
}

@techreport{lewis1997reuters,
  author      = {Lewis, David D.},
  title       = {Reuters-21578 Text Categorization Test Collection},
  institution = {AT\&T Labs Research},
  year        = {1997}
}

@article{debole2005analysis,
  author    = {Debole, Franca and Sebastiani, Fabrizio},
  title     = {An Analysis of the Relative Hardness of Reuters-21578 Subsets},
  journal   = {Journal of the American Society for Information Science and Technology},
  volume    = {56},
  number    = {6},
  pages     = {584--596},
  year      = {2005}
}

@software{labelfusion2025,
  title = {LabelFusion: Learning to Fuse LLMs and Transformer Classifiers for Robust Text Classification},
  author = {{Anonymous}},
  year = {2025},
  url = {https://anonymous.4open.science/r/}
}

\end{document}